\documentclass{article}

\usepackage[main, final]{neurips_2025}

\usepackage[utf8]{inputenc} 
\usepackage[T1]{fontenc}    
\usepackage{hyperref}       
\usepackage{url}            
\usepackage{booktabs}       
\usepackage{amsfonts}       
\usepackage{nicefrac}       
\usepackage{microtype}      
\usepackage{xcolor}         
\usepackage{graphicx}       
\usepackage{amsmath}        
\usepackage{subcaption}

\title{Conditional Flow Matching for Continuous Anomaly Detection in Autonomous Driving on a Manifold-Aware Spectral Space}

\author{%
  Antonio Guillen-Perez  \\
  Independent Researcher \\
  \texttt{antonio\_algaida@hotmail.com} \\
  \href{https://antonioalgaida.github.io/}{antonioalgaida.github.io}
}

\begin{document}

\maketitle

\begin{abstract}
Safety validation for Level 4 autonomous vehicles (AVs) is currently bottlenecked by the inability to scale the detection of rare, high-risk ``long-tail'' scenarios using traditional rule-based heuristics. We present \textbf{Deep-Flow}, an unsupervised framework for safety-critical anomaly detection that utilizes \textbf{Optimal Transport Conditional Flow Matching} (OT-CFM) to characterize the continuous probability density of expert human driving behavior. Unlike standard generative approaches that operate in unstable, high-dimensional coordinate spaces, Deep-Flow constrains the generative process to a \textbf{low-rank spectral manifold} via a Principal Component Analysis (PCA) bottleneck. This ensures kinematic smoothness by design and enables the computation of the \textbf{exact Jacobian trace} for numerically stable, deterministic log-likelihood estimation. To resolve multi-modal ambiguity at complex junctions, we utilize an \textbf{Early Fusion Transformer} encoder with \textbf{lane-aware goal conditioning}, featuring a direct skip-connection to the flow head to maintain intent-integrity throughout the network. Furthermore, we introduce a \textbf{kinematic complexity weighting} scheme that prioritizes high-energy maneuvers (quantified via path tortuosity and jerk) during the simulation-free training process. Evaluated on the \textbf{Waymo Open Motion Dataset} (WOMD), our framework achieves an \textbf{AUC-ROC of 0.766} against a heuristic golden set of safety-critical events. More significantly, our analysis reveals a fundamental distinction between kinematic danger and \textbf{semantic non-compliance}. Deep-Flow identifies a critical ``predictability gap'' by surfacing out-of-distribution behaviors, such as lane-boundary violations and non-normative junction maneuvers, that traditional safety filters overlook. This work provides a mathematically rigorous foundation for defining \textbf{statistical safety gates}, enabling objective, data-driven validation for the safe deployment of autonomous fleets. Code and pre-trained checkpoints are available at \url{https://github.com/AntonioAlgaida/FlowMatchingTrajectoryAnomaly}
\end{abstract}

\section{Introduction}
\label{sec:intro}

The deployment of Level 4 (L4) autonomous vehicles (AVs) hinges on the ability to provide rigorous safety argumentation for complex, high-dimensional operational design domains (ODDs). Traditional validation strategies, largely dependent on aggregate metrics such as miles per disengagement, are increasingly recognized as insufficient for capturing the ``long-tail'' of safety-critical events. To achieve the reliability required for commercial scale, the industry must transition toward automated, data-driven methods for surfacing ``unknown unknowns'', scenarios that are kinematically feasible but semantically or socially out-of-distribution (OOD).

\begin{figure}[t]
    \centering
    \includegraphics[width=\textwidth]{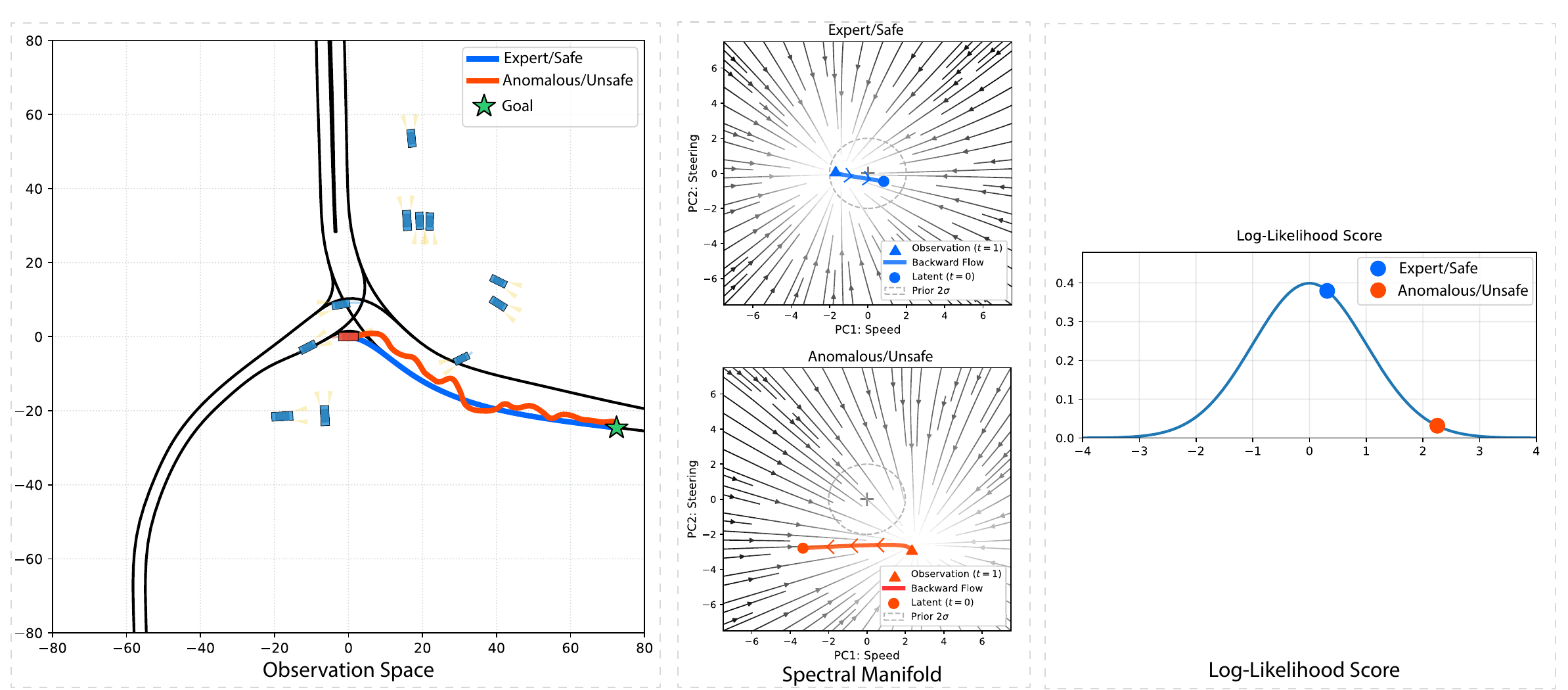}
    \caption{\textbf{Overview of the Deep-Flow Framework.} (Left) We observe an agent's trajectory within a goal-conditioned context. While both safe (Blue) and anomalous (Orange) maneuvers may reach the same goal, they represent different densities on the driving manifold. (Center) Trajectories are projected into a low-rank spectral manifold where backward ODE integration ($t=1 \to 0$) maps maneuvers to a Gaussian prior. (Right) Deep-Flow identifies safety-critical anomalies by mapping non-normative behaviors to the low-probability tails of the expert distribution, providing a continuous and mathematically rigorous safety score.}
    \label{fig:teaser}
\end{figure}

Current safety validation pipelines rely heavily on rule-based heuristics, such as longitudinal deceleration thresholds (e.g., $a < -5.0 \, m/s^2$) or Time-to-Collision (TTC) triggers. While effective for identifying obvious kinetic hazards, these methods are fundamentally brittle. They are blind to \textit{semantic anomalies}, such as lane-boundary violations, illegal maneuvers, or aggressive social interactions that do not involve extreme braking. Furthermore, supervised learning approaches for anomaly detection are limited by the extreme scarcity of labeled incident data. This motivates a shift toward \textit{unsupervised generative modeling}, where a system learns the continuous probability density function of expert human behavior. Under this paradigm, as illustrated in \textbf{Fig. \ref{fig:teaser}}, a safety violation is defined as a statistically rare deviation from the learned expert manifold.

Existing generative architectures present significant trade-offs for safety validation. Autoregressive (AR) models, such as MotionLM \cite{seff2023motionlm}, are prone to exposure bias and temporal drift over long planning horizons (e.g., 8 seconds), which can pollute the resulting likelihood scores with numerical artifacts. Conversely, while Diffusion models generate high-fidelity samples, they rely on stochastic differential equations that make exact log-likelihood estimation computationally prohibitive for large-scale fleet auditing. Variational Autoencoders (VAEs) offer tractable likelihoods but often suffer from \textit{posterior collapse}, resulting in blurry, unimodal distributions that fail to resolve the complex multi-modality of urban driving.

We propose \textbf{Deep-Flow}, a holistic generative framework for anomaly detection based on \textbf{Optimal Transport Conditional Flow Matching} (OT-CFM) \cite{Tong2023Feb}. Deep-Flow addresses the limitations of coordinate-space modeling by operating on a low-rank \textbf{Spectral Manifold}. By projecting trajectories into a whitened PCA coefficient space, we enforce kinematic smoothness by design and enable the stable computation of the exact Jacobian trace for deterministic likelihood estimation. To the best of our knowledge, this is the first framework to combine manifold-aware flow matching with topologically-grounded goal conditioning for AV safety validation.

The primary contributions of this work are as follows:
\begin{itemize}
    \item \textbf{Spectral Manifold Bottleneck:} We demonstrate that regressing spectral coefficients rather than raw waypoints acts as a rigorous low-pass filter, ensuring that generated trajectories remain kinematically feasible and numerically stable for likelihood integration.
    \item \textbf{Lane-Aware Goal Conditioning:} We introduce a skip-connection architecture that injects topological lane geometry directly into the flow head, resolving the multi-modal ambiguity inherent in complex junctions and roundabouts.
    \item \textbf{Kinematic Complexity Weighting:} We propose a physics-informed importance sampling scheme based on path tortuosity and jerk energy, forcing the model to prioritize the learning of high-energy, safety-relevant maneuvers over routine cruising behavior.
    \item \textbf{Semantic Anomaly Discovery:} We evaluate our framework on the Waymo Open Motion Dataset (WOMD), achieving an AUC-ROC of 0.766 and proving that Deep-Flow identifies a critical ``predictability gap'' by surfacing semantic violations that rule-based heuristics overlook.
\end{itemize}

\section{Related Work}
\label{sec:related_work}

The validation of autonomous systems has evolved from deterministic replay to probabilistic density estimation. Our work situates itself at the intersection of high-fidelity motion forecasting, continuous generative modeling, and unsupervised OOD detection.

\subsection{State-of-the-Art Motion Forecasting}
Modern trajectory prediction has transitioned from anchor-based heuristics to scene-centric transformer architectures. Early successes such as \textit{VectorNet} \cite{gao2020vectornet} established the efficacy of vectorized scene representations. Recently, the field has been dominated by factorized attention mechanisms. For example, \textit{Wayformer} \cite{nayakanti2023wayformer} demonstrated the efficiency of early-fusion strategies, while the \textit{Motion Transformer (MTR)} \cite{shi2022motion} and \textit{QCNet} \cite{zhou2023query} utilized query-centric designs to capture the multi-modal nature of human intent.

However, these discriminative models are fundamentally optimized for \textit{mode-seeking} (Accuracy) rather than \textit{density estimation} (Safety). As discussed in our previous work on offline policy learning \cite{guillen2025imitation}, purely imitation-based objectives often yield policies that are brittle in closed-loop execution. While Offline RL methods, such as Conservative Q-Learning (CQL), can mitigate this via value-function regularization \cite{guillen2025mining}, they typically require extensive reward engineering. Deep-Flow addresses this by learning a multi-modal continuous safety manifold directly from expert data, providing a self-supervised density signal without manual reward specification.

\subsection{Generative Modeling: From Diffusion to Flows}
Generative modeling in robotics has seen a rapid paradigm shift. \textit{Variational Autoencoders} (VAEs) and \textit{Generative Adversarial Networks} (GANs) effectively pioneered stochastic prediction but suffered from \textit{posterior collapse} and training instability, respectively. The introduction of \textit{Denoising Diffusion Probabilistic Models} (DDPMs) \cite{ho2020denoising} and \textit{Latent Diffusion Models} \cite{guillen2025virtuoso} resolved these issues, enabling high-fidelity trajectory generation as seen in \textit{MotionDiffuser} \cite{jiang2023motiondiffuser} and \textit{Guided Diffusion} for planning \cite{janner2022planning}.

However, Diffusion models rely on Stochastic Differential Equations (SDEs) that require iterative denoising, making the computation of \textbf{exact log-likelihoods} intractable for large-scale fleet auditing. Furthermore, stochastic trajectories in diffusion often lack kinematic smoothness without heavy post-processing. \textit{Conditional Flow Matching} (CFM) \cite{Lipman2022Oct, Tong2023Feb} and \textit{Rectified Flows} \cite{liu2022flow} have emerged as a superior alternative, defining a deterministic Ordinary Differential Equation (ODE) that maps noise to data in a one-to-one way. By utilizing \textbf{Optimal Transport (OT)} paths, CFM ensures straight-line trajectories in the probability flow, significantly reducing the numerical ``stiffness'' of the ODE solver. Deep-Flow leverages this property to enable \textit{exact likelihood computation} via the instantaneous change of variables formula, a capability largely absent in current diffusion-based AV baselines.

\subsection{Autoregressive vs. Holistic Representations}
A prominent industry direction, exemplified by \textit{MotionLM} \cite{seff2023motionlm} and \textit{DriveGPT} \cite{Huang2024Dec}, frames driving as a discrete token-sequence modeling task. These Autoregressive (AR) models leverage the scaling laws of Large Language Models (LLMs) to predict discretized waypoints.

While AR models scale effectively, they fundamentally suffer from \textbf{exposure bias}. During inference, microscopic errors accumulate temporally, leading to ``hallucinated physics'' over long horizons (e.g., 8s). For safety validation, this temporal drift renders likelihood scores unreliable, as the joint probability of a trajectory tends to vanish as $T \to \infty$. Deep-Flow avoids this by adopting a \textbf{Holistic} representation. By treating the trajectory as a single high-dimensional primitive on a spectral manifold (Sec \ref{sec:spectral_manifold}), we avoid the product-rule instability of sequence models and ensure global kinematic consistency by design.

\subsection{Unsupervised Anomaly Detection in Driving}
Anomaly detection in AVs is typically divided into \textit{rule-based} and \textit{reconstruction-based} methods. Rule-based systems rely on predefined kinematic thresholds (TTC, jerk limits), which fail to capture semantic anomalies such as wrong-way driving. Reconstruction-based methods utilize Autoencoders to detect anomalies via high reconstruction error \cite{bolte2019anomaly}. However, these models often suffer from the \textbf{generalization trap}, where they successfully reconstruct simple anomalies, thereby failing to flag them. 

Semantic-front methods, such as the neuro-symbolic \textit{Semantic-Drive} framework \cite{guillen2025semantic}, where we utilize Vision-Language Models (VLMs) to ground open-vocabulary queries (e.g., ``erratic jaywalking''). While effective for known categories, these methods are limited by textual descriptiveness. Deep-Flow operates as a pure \textbf{One-Class Classifier}, learning the manifold of expert driving exclusively. By utilizing the \textbf{Jacobian trace} of the flow field, we provide a mathematically rigorous density estimate that serves as a continuous, physics-aware safety gate.

\section{Methodology}
\label{sec:meth}

We propose \textbf{Deep-Flow}, an unsupervised generative framework designed for safety-critical anomaly detection in autonomous driving. Deep-Flow models the conditional probability density of expert human behavior as a \textit{Continuous Normalizing Flow} (CNF). Unlike traditional discriminative approaches that require labeled incident data, Deep-Flow learns the high-dimensional manifold of nominal behavior and identifies anomalies as statistically rare events. 

The architecture is composed of three interconnected modules: (i) a \textit{Goal-Conditioned Early Fusion Encoder} for high-fidelity scene understanding; (ii) a \textit{Spectral Manifold Bottleneck} that enforces kinematic feasibility; and (iii) a \textit{Conditional Flow Matching} (CFM) head used for exact likelihood estimation.

\subsection{Problem Formulation}
We frame safety validation as an unsupervised OOD detection task. Let $\mathbf{x} \in \mathbb{R}^{D}$ represent an agent's future trajectory over a fixed planning horizon (where $D = T \times 2$), and let $\mathbf{C}$ represent the spatio-temporal scene context, including map topology, multi-agent history, and dynamic traffic light states. Our objective is to estimate the conditional probability density function $p(\mathbf{x} \mid \mathbf{C})$ derived from a large-scale dataset of expert human demonstrations $\mathcal{D} = \{(\mathbf{x}_i, \mathbf{C}_i)\}_{i=1}^N$.

Following the principles of probabilistic safety validation, we define the \textbf{anomaly score} $\mathcal{A}$ as the negative log-likelihood (NLL) of an observed trajectory $\mathbf{x}_{\text{obs}}$ conditioned on its environment:
\begin{equation}
    \mathcal{A}(\mathbf{x}_{\text{obs}}, \mathbf{C}) = -\log p_\theta(\mathbf{x}_{\text{obs}} \mid \mathbf{C})
\end{equation}
In this formulation, $\mathcal{A}$ serves as a continuous, mathematically rigorous proxy for safety risk. High-likelihood regions correspond to maneuvers well-represented in the expert nominal distribution. Conversely, the low-probability ``tails'' of the distribution identify safety-critical events (such as near-misses, geometric violations, or erratic control) that deviate significantly from learned human norms.

\textbf{Holistic vs. Autoregressive Modeling.} State-of-the-art motion forecasting often relies on AR formulations \cite{seff2023motionlm, zhang2023trafficbots, mao2023gpt}, which decompose the joint density into a product of step-wise conditionals: $p(\mathbf{x} \mid \mathbf{C}) = \prod_{t=1}^T p(x_t \mid x_{<t}, \mathbf{C})$. While effective for sequence prediction, AR approaches are inherently susceptible to \textit{exposure bias} and \textit{temporal drift}, where microscopic errors in the early steps of the horizon compound into macroscopic deviations.

For safety validation, a drift of even one meter over an 8-second horizon can be the difference between a safe maneuver and a collision, making AR likelihoods noisy for OOD thresholding. Furthermore, the cumulative product of 80+ probabilities often leads to \textit{vanishing likelihoods} and numerical instability. In contrast, Deep-Flow adopts a \textbf{holistic generative approach} via Flow Matching. By treating the entire trajectory as a single primitive on a low-dimensional manifold, we ensure global kinematic consistency and leverage the \textit{instantaneous change of variables} formula (more information on \ref{sec:inference}). This allows us to compute an exact, numerically stable log-likelihood scalar that integrates the density along a continuous ODE path, providing a reliable signal for detecting complex, long-tail anomalies.

\subsection{The Spectral Manifold Representation}
\label{sec:spectral_manifold}

A significant challenge in high-dimensional trajectory modeling ($ \mathbf{x} \in \mathbb{R}^{160} $ for an 8-second horizon at 10Hz) is the presence of high-frequency artifacts and kinematically infeasible "jitter." While standard regression models rely on auxiliary loss terms to penalize jerk, we propose to enforce kinematic feasibility \textit{by design} through a \textbf{Spectral Manifold Bottleneck}.

Human driving behavior is intrinsically low-rank, constrained by non-holonomic \footnote{Nonholonomic systems have constraints on their velocities that cannot be reduced to constraints on position. For example, a car cannot move sideways; its motion is restricted to directions aligned with its heading.} vehicle dynamics and second-order smoothness requirements. To exploit this structure, we project raw trajectory coordinates into a \textbf{Spectral Coefficient Space} using Principal Component Analysis (PCA). We derive an orthogonal basis $\mathbf{B} \in \mathbb{R}^{k \times 160}$ and a population mean $\mu$ from the expert demonstration set. Any observed or generated trajectory $\mathbf{x}$ is represented by a latent vector $\mathbf{z} \in \mathbb{R}^k$:
\begin{equation}
    \mathbf{z} = \mathbf{W}^{-1} \mathbf{B}^T (\mathbf{x} - \mu)
\end{equation}
where $\mathbf{W} = \text{diag}(\sigma_1, \dots, \sigma_k)$ is a \textit{whitening matrix} containing the standard deviations of the principal components. This transformation serves three critical architectural purposes:

\begin{enumerate}
    \item \textbf{Implicit Kinematic Regularization:} By selecting $k=12$ components, we capture $>99\%$ of the dataset variance while effectively filtering out high-frequency sensor noise and measurement jitter. The top eigenvectors (``eigen-trajectories'') represent smooth geometric primitives (such as constant-velocity progress or sustained lateral curvature) ensuring that any linear combination remains $C^2$ continuous. A visual representation of what each PC component captures can be seen in Appendix \ref{app:pca_analysis}.
    \item \textbf{Manifold Whitening:} Normalizing by $\mathbf{W}$ ensures the target data distribution $p_1$ is nearly isotropic ($ \mathcal{N}(0, \mathbf{I}) $). This minimizes the \textit{transport cost} in our Flow Matching objective by aligning the scale of the data manifold with the Gaussian prior $p_0$, leading to significantly more stable vector field convergence.
    \item \textbf{Computational Tractability:} Reducing the dimensionality from 160 to 12 allows for the computation of the \textbf{exact Jacobian trace} during inference (see Sec.~\ref{sec:inference}), since the computational cost grows with the latent dimensionality $k$. With a small $k$, exact evaluation becomes tractable, avoiding the need for high-variance stochastic trace estimators (e.g., Hutchinson's) required for high-dimensional flows, providing a precise and deterministic anomaly score.
\end{enumerate}

By shifting the generative task from predicting unstable waypoints to regressing stable \textit{maneuver coefficients}, Deep-Flow focuses its capacity on the semantic intent of the agent rather than the microscopic details of the path.

\begin{figure}[t]
    \centering
    \hspace{0.2\textwidth}
    \includegraphics[width=0.75\textwidth]{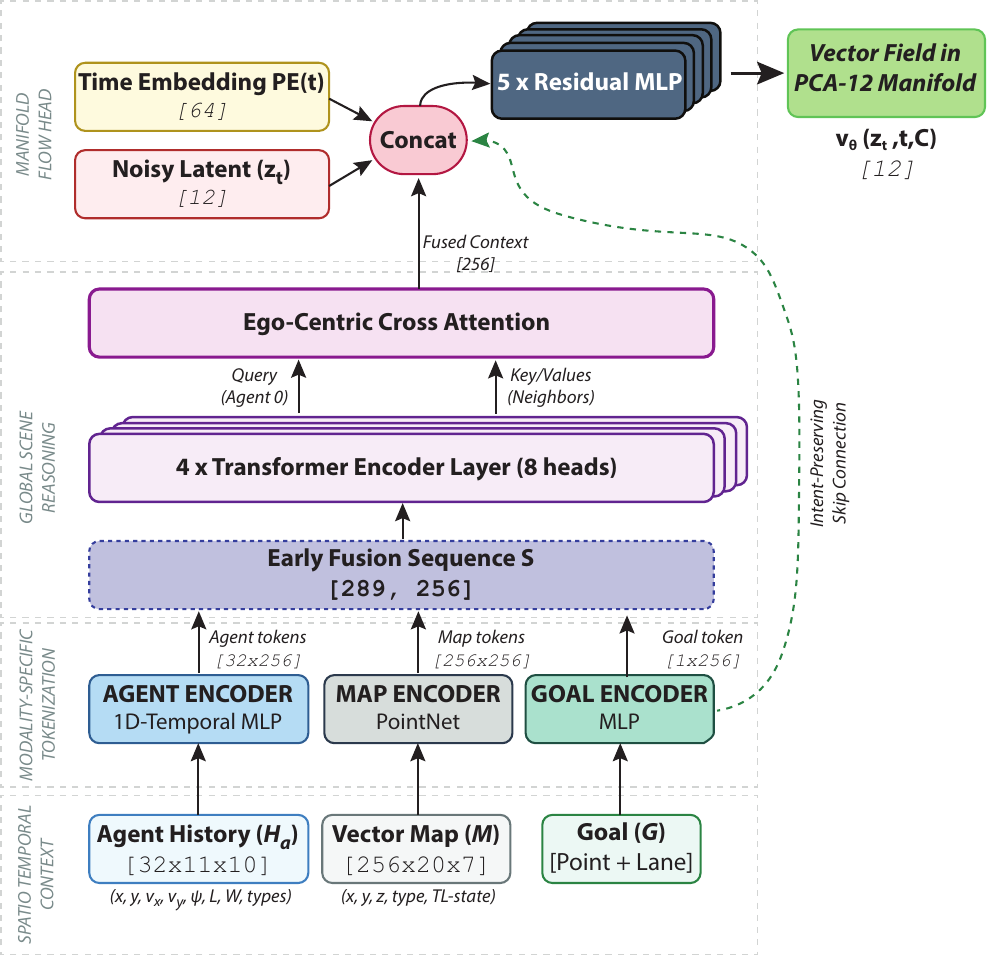}
    \caption{\textbf{Deep-Flow Encoder Architecture.} Heterogeneous modalities are tokenized and fused via a Hierarchical Transformer. The Goal signal is injected twice: once in the global context and once as a direct skip-connection to the Flow Head to preserve intent-integrity.}
    \label{fig:architecture}
\end{figure}

\subsection{Goal-Conditioned Scene Encoding}
\label{sec:encoder}

To process the heterogeneous spatio-temporal modalities, specifically dynamic agent histories and static vectorized map elements, we adopt an \textbf{Early Fusion} transformer architecture, as described in \cite{nayakanti2023wayformer}. While hierarchical fusion strategies provide modality-specific processing, Early Fusion enables the model to learn complex cross-modal dependencies (e.g., the influence of a specific stop-line on an agent's deceleration profile) at the earliest stages of the network, which is critical for high-fidelity density estimation.

\textbf{Goal-Lane Conditioning.} A primary failure mode in probabilistic trajectory modeling is multi-modal ambiguity at decision points, such as intersections. To resolve this for safety validation, we explicitly condition the probability density on a \textit{semantic intent}. We extract the lane centerline polyline closest to the agent's final observed position, denoted as $\mathbf{g}_{\text{lane}} \in \mathbb{R}^{20 \times 2}$. This topological feature serves as a ``guide rail'' for the generative flow, narrowing the manifold to behaviors consistent with a specific navigation goal.

The encoder architecture (Fig. \ref{fig:architecture}) proceeds through four distinct stages:

\begin{enumerate}
    \item \textbf{Modality-Specific Tokenization:} We use distinct MLP-based encoders to project raw features into a shared latent space $D=256$. For agents, the temporal history is flattened into a single social token to capture kinematic trends. For map polylines, we utilize a symmetric \textit{PointNet}-style encoder (Linear layer followed by Max-Pooling) to ensure \textbf{permutation invariance} across the points of the polyline, making the representation robust to the direction of lane indexing.
    
    \item \textbf{Global Context Fusion:} The Modality tokens are concatenated with a learned Goal Embedding to form a unified sequence $S \in \mathbb{R}^{(N_{\text{agents}} + N_{\text{map}} + 1) \times D}$. This sequence is processed by a 4-layer Transformer Encoder using \textit{Pre-Norm} blocks and \textit{GELU} activations for improved gradient stability.
    
    \item \textbf{Ego-Centric Cross-Attention:} Rather than using a global average of the scene, we utilize a \textbf{Cross-Attention} mechanism where the Ego-agent token ($Agent$ $0$) acts as the \textit{Query} to pool relevant features from the global sequence. Unlike full self-attention, which weights all interactions equally, this query-based approach allows the model to perform \textit{spatial filtering}. It forces the network to ignore distant, irrelevant agents and focus capacity on the specific map topology and neighbors that physically constrain the Ego's manifold.
    
    \item \textbf{The Direct Intent Skip-Connection:} In deep transformers, conditioning signals (like the Goal) can suffer from \textit{signal dilution} as they pass through multiple attention layers. To maintain the integrity of the navigation target, the Goal Embedding is bypassed via a \textbf{direct skip-connection} and concatenated to the Transformer output before entering the Flow Head. This ensures that the learned vector field $\mathbf{v}_\theta$ is explicitly and strongly anchored to the goal coordinates, preventing mode dispersion and ensuring the flow converges tightly on the intended destination.
\end{enumerate}

\subsection{Optimal Transport Conditional Flow Matching}
\label{sec:cfm}

We utilize \textit{Conditional Flow Matching} (CFM) \cite{Lipman2022Oct, Tong2023Feb} to learn a time-dependent vector field $\mathbf{v}_\theta(\mathbf{z}, t, \mathbf{C})$ that defines a bijective mapping between a simple Gaussian prior $p_0 = \mathcal{N}(0, \mathbf{I})$ and the complex distribution of expert spectral coefficients $p_1$. Unlike Diffusion models that rely on stochastic differential equations (SDEs), CFM enables \textbf{simulation-free training} of Continuous Normalizing Flows (CNFs), leading to more stable convergence and highly efficient inference.

\textbf{Probability Paths and Optimal Transport.} We define a conditional probability path $p_t(\mathbf{z})$ that interpolates between noise and data. To ensure the most efficient transport of probability mass, we adopt the \textbf{Optimal Transport (OT)} displacement map. This path corresponds to the $W_2$ (2-Wasserstein\footnote{The 2-Wasserstein distance measures the minimum expected squared Euclidean distance required to transport mass from one distribution to another; its geodesics correspond to linear interpolation between samples.}) geodesic, which defines a linear, constant-velocity trajectory between a noise sample $\mathbf{z}_0$ and an expert sample $\mathbf{z}_1$:
\begin{equation}
    \psi_t(\mathbf{z}) = (1 - (1 - \sigma_{\min})t)\mathbf{z}_0 + t \mathbf{z}_1
\end{equation}
where $\sigma_{\min} = 10^{-4}$ is a small stability constant that ensures the conditional density remains well-defined at $t=0$. 

\textbf{Vector Field Regression.} The corresponding target velocity field $\mathbf{u}_t(\mathbf{z} \mid \mathbf{z}_1)$ for this OT path is notably simple and time-independent: $\mathbf{u}_t(\mathbf{z} \mid \mathbf{z}_1) = \mathbf{z}_1 - (1-\sigma_{\min})\mathbf{z}_0$. The generative model $\mathbf{v}_\theta$ is trained to regress this target velocity conditioned on the scene context $\mathbf{C}$ (from Sec \ref{sec:encoder}). The training objective is formulated as the expectation over the path:
\begin{equation}
\label{eq:cfm}
    \mathcal{L}_{\text{CFM}} = \mathbb{E}_{t \sim \mathcal{U}[0,1], \mathbf{z}_0 \sim p_0, \mathbf{z}_1 \sim p_1} \left[ \left\| \mathbf{v}_\theta(\psi_t(\mathbf{z}_0), t, \mathbf{C}) - \left( \mathbf{z}_1 - (1-\sigma_{\min})\mathbf{z}_0 \right) \right\|^2 \right]
\end{equation}

By minimizing this objective, $\mathbf{v}_\theta$ learns to approximate the score-front of the expert distribution. The \textbf{straight-line trajectories} inherent to OT-CFM provide a significant engineering advantage: they are easier for ODE solvers to integrate than the curved paths produced by standard Diffusion or Flow-based methods. This reduced curvature results in a more stable Jacobian\footnote{The Jacobian of the transformation describes how local volumes change under the flow, and its trace determines the change in log-density along the trajectory.} during backward integration, which is paramount for the numerical accuracy of our log-likelihood safety metric.

\subsection{Kinematic Complexity and Manifold Grounding}
\label{sec:loss}

A persistent failure mode in large-scale trajectory learning is the dominance of nominal, constant-velocity regimes (e.g., highway cruising), which causes the model to underfit the ``long-tail'' of safety-critical maneuvers. To mitigate this dataset imbalance and ensure high-fidelity learning of rare maneuvers, we propose a \textbf{Kinematic Complexity Weighting} scheme combined with a \textbf{Hybrid Manifold-Physical} loss.

\textbf{Importance Sampling via Differential Geometry.} We dynamically weight the contribution of each sample $i$ to the gradient based on its geometric and dynamic complexity. Drawing inspiration from \textit{importance sampling} strategies used to handle imbalanced behavioral distributions \cite{Bansal2018Dec, chai2021multipath}, we define a complexity weight $w_i$:
\begin{equation}
    w_i = \underbrace{\left( \frac{\int_0^T \|\dot{\mathbf{x}}_t\| dt}{\|\mathbf{x}_T - \mathbf{x}_0\|} \right)}_{\text{Tortuosity } \tau} \times \underbrace{\left( \text{exp} \left( \alpha \int_0^T \|\dddot{\mathbf{x}}_t\|^2 dt \right) \right)}_{\text{Jerk Energy } \mathcal{J}}
\end{equation}
where $\tau$ is the ratio of path length to net displacement (identifying non-linear maneuvers like roundabouts or U-turns) and $\mathcal{J}$ is the integrated squared jerk (identifying high-energy interactions such as emergency braking or evasive swerving). This weighting forces the vector field $\mathbf{v}_\theta$ to prioritize the nuanced control inputs required for rare maneuvers over the trivial interpolation of straight-line driving.

\textbf{Euclidean Grounding.} While the CFM objective (Eq. \ref{eq:cfm}) operates in the spectral coefficient space $\mathbf{z} \in \mathbb{R}^k$, small numerical errors in the latent manifold can translate into significant Average Displacement Error (ADE) in the physical world, particularly at the trajectory horizons. To ensure the learned manifold remains \textit{physically grounded}, we introduce a secondary reconstruction loss in Euclidean coordinates. 

We map the predicted latent velocity $\mathbf{v}_\theta$ back to the coordinate space using the fixed PCA basis $\mathbf{B}$ and mean $\mu$ (defined in Sec \ref{sec:spectral_manifold}). The total multi-objective loss is formulated as:
\begin{equation}
    \mathcal{L}_{\text{total}} = \frac{1}{B} \sum_{i=1}^B \bar{w}_i \left( \mathcal{L}_{\text{CFM}} + \lambda \sqrt{\frac{1}{T} \sum_{t=1}^T \|\hat{\mathbf{x}}_{i,t} - \mathbf{x}_{i,t}\|^2} \right)
\end{equation}
where $\bar{w}_i$ are batch-normalized weights to prevent gradient explosion, and $\lambda$ is a balancing coefficient. We specifically utilize the \textbf{Root Mean Squared Error (RMSE)} for the coordinate term to maintain linear units (meters) relative to the quadratic flow loss, preventing either term from dominating the optimization landscape. This hybrid approach ensures that Deep-Flow is not only probabilistically sound in the latent space but also geometrically precise in the real-world ODD.

\subsection{Likelihood Estimation and Inference}
\label{sec:inference}

To transform the learned vector field $\mathbf{v}_\theta$ into a safety-critical metric, we perform unsupervised anomaly detection by computing the \textit{exact log-likelihood} of observed trajectories. Unlike discrete normalizing flows that rely on the determinant of a Jacobian, CNFs allow us to evaluate the probability density by integrating the local expansion and contraction of the manifold along the flow.

\textbf{Probability Mass Conservation and the Continuity Equation.} The evolution of a probability density $p(\mathbf{z}, t)$ under a vector field $\mathbf{v}$ is governed by the \textit{continuity equation} (or Liouville's equation in statistical mechanics):
\begin{equation}
    \frac{\partial p(\mathbf{z}, t)}{\partial t} = -\text{div}(p(\mathbf{z}, t)\mathbf{v})
\end{equation}
By applying the log-derivative trick, we obtain the \textbf{instantaneous change of variables formula}, which defines the rate of change of the log-probability for a particle $\mathbf{z}_t$ moving along the ODE:
\begin{equation}
    \frac{d \log p(\mathbf{z}_t, t)}{dt} = -\text{Tr} \left( \frac{\partial \mathbf{v}_\theta(\mathbf{z}_t, t, \mathbf{C})}{\partial \mathbf{z}_t} \right)
\end{equation}

where the \textit{Trace of the Jacobian} ($\text{Tr}(\mathbf{J})$) represents the divergence of the flow. Intuitively, the divergence measures how the vector field locally expands or contracts the space as trajectories evolve over time.

A \textbf{positive divergence} indicates that nearby trajectories are moving away from each other, meaning that the flow is \textit{expanding} the volume around that region. As the same probability mass is distributed over a larger volume, the local density decreases, resulting in a lower likelihood for any observation passing through that region. In other words, the model considers such regions less probable under the learned data distribution.

In contrast, a \textbf{negative divergence} indicates that trajectories are converging, meaning that the flow is \textit{contracting} the space. This concentrates probability mass into a smaller volume, increasing the local density and therefore the likelihood of trajectories that pass through that region.

From a dynamical perspective, regions of \textbf{contraction} correspond to \textit{high-density manifolds} learned from the data, where typical trajectories tend to lie, while regions of \textbf{expansion} correspond to \textit{low-density or uncertain areas} that the model has not frequently observed. As a result, trajectories that consistently pass through expanding regions accumulate a lower overall likelihood and are identified as anomalous.

\textbf{Exact Trace Computation.} A critical advantage of our \textit{Spectral Manifold Bottleneck} (Sec. \ref{sec:spectral_manifold}) is the reduction of the latent space to $k=12$ dimensions. In high-dimensional flows, computing the divergence term is computationally prohibitive, typically requiring stochastic approximations like \textit{Hutchinson’s Trace Estimator} \cite{hutchinson1989stochastic}, which introduces sampling variance into the anomaly score. Because Deep-Flow operates in a low-rank subspace ($k=12$), we can compute the \textbf{exact Jacobian trace} efficiently using automatic differentiation:
\begin{equation}
    \text{Tr}(\nabla_\mathbf{z} \mathbf{v}_\theta) = \sum_{j=1}^{k} \frac{\partial v_{\theta, j}}{\partial z_{t, j}}
\end{equation}
We evaluate the total log-likelihood by solving the ODE backwards from the observation $\mathbf{z}_1$ ($t=1$) to the prior noise $\mathbf{z}_0$ ($t=0$):
\begin{equation}
    \log p(\mathbf{z}_1 \mid \mathbf{C}) = \log p_0(\mathbf{z}_0) - \int_{0}^{1} \text{Tr} \left( \nabla_{\mathbf{z}_t} \mathbf{v}_\theta \right) dt
\end{equation}
We implement this via a fixed-step Runge-Kutta 4 (RK4) \cite{rungekutta} integrator. The resulting log-likelihood is deterministic and numerically stable, providing a high-fidelity signal for the detection of OOD behaviors. Scenarios that land in the tails of the Gaussian prior $p_0$ or traverse regions of high flow divergence are assigned a high \textit{anomaly score} $\mathcal{A} = -\log p(\mathbf{z}_1 \mid \mathbf{C})$.

\section{Experimental Setup}
\label{sec:experiments}

We evaluate \textbf{Deep-Flow} using the \textbf{Waymo Open Motion Dataset (WOMD)} \cite{ettinger2021waymo} to verify its efficacy in characterizing safety-critical anomalies. Our experimental design prioritizes high-fidelity scene representation and rigorous hardware-aware optimization to ensure scalable training and deterministic inference.

\subsection{Data Curation and Representation}
WOMD provides a diverse corpus of urban and suburban driving logs. From the raw protobuf segments, we extract a training set of 250,000 scenarios and a validation set of 8,856 scenarios, maintaining a consistent sampling frequency of 10Hz.

\textbf{Ego-Centric Normalization:} To ensure the learned vector field is translation and rotation invariant, all scenarios are transformed into a \textit{canonical ego-centric frame} at the anchor time $t=10$ (1.1s into the log). The ego-vehicle is positioned at $(0,0)$ facing the positive x-axis. All coordinates are normalized by a fixed scale factor of 50.0 meters to maintain numerical stability during ODE integration.

\textbf{Unified Multi-Modal Representation.} The scene context $\mathbf{C}$ is represented as a spatio-temporal graph:
\begin{itemize}
    \item \textbf{Dynamic Agents:} We select the 32 nearest neighbors relative to the ego-vehicle. Features include 2D position, velocity, heading, object dimensions, and type of agent (vehicle, bicycle, pedestrian, etc.)
    \item \textbf{Vectorized Map:} We query the 256 nearest map polylines (lane centers, road edges, and crosswalks), each subsampled to 20 vertices. 
    \item \textbf{Signal-to-Geometry Binding:} Unlike models that treat traffic lights as global features, we perform \textit{spatial binding} by appending the one-hot encoded signal state (Red, Yellow, Green) directly to the coordinate features of the lane center polyline it controls. This provides a local, topologically grounded context for regulatory compliance.
\end{itemize}

\subsection{The ``Golden Test Set'' for Safety Validation}
A primary challenge in unsupervised anomaly detection is the lack of explicit labels for the ``long-tail.'' To quantitatively evaluate Deep-Flow, we curate a \textbf{Golden Test Set} of known safety-critical events by mining the validation logs using two high-confidence kinematic heuristics:
\begin{enumerate}
    \item \textbf{Extreme Deceleration:} Any scenario where the Ego-vehicle or a primary agent exhibits a longitudinal acceleration $a < -5.0 \, m/s^2$, signaling emergency braking.
    \item \textbf{Dynamic Instability:} Any scenario where the yaw rate exceeds $1.5 \, rad/s$, indicating a sudden swerve or loss of control.
\end{enumerate}
This heuristically-mined subset serves as the ground-truth positive class for our AUC-ROC evaluation, allowing us to quantify the alignment between the learned probability density and physical hazard metrics.

\subsection{Implementation Details}
Deep-Flow is implemented in PyTorch and trained on a single NVIDIA GeForce RTX 3090 (24GB VRAM), Intel i5-13600K, and 64GB RAM. 
\textbf{Network Architecture:} The Scene Encoder utilizes a 4-layer Transformer with 8 attention heads and a hidden dimension of 256. The Flow Head is a Residual MLP with 5 blocks and a hidden dimension of 1024. We found that increasing the head width was critical to handle the high-information density of the skip-conditioned goal lane.

\textbf{Training Protocol:} We train the model for 80 epochs using the \textbf{AdamW} optimizer with a batch size of 256, optimized for 24GB of VRAM. We employ a \textbf{Cosine Annealing} learning rate scheduler starting at $5 \times 10^{-4}$ with a weight decay of $10^{-2}$. 

\textbf{Numerical Stability:} To prevent gradient explosions common in Flow Matching and Transformers, we apply global gradient clipping at a norm of 1.0. All experiments are conducted in FP32 precision to ensure the accuracy of the Jacobian trace calculations.

\textbf{Engineering Optimizations:} To maximize GPU utilization and bypass the WSL2 file-system bottleneck, we implemented a \textbf{Parallel Eager Loader}. This system pre-processes and caches the dataset into system RAM (64GB) as a list of lightweight NumPy tuples during initialization. This reduced epoch time from 15 minutes to under 3 minutes, maintaining $>95\%$ GPU utilization throughout the training run.

\textbf{Inference:} Likelihood estimation is performed using a fixed-step Runge-Kutta 4 (RK4) integrator with 20 steps. We calculate the exact trace of the Jacobian for the 12-dimensional manifold, ensuring a precise and deterministic anomaly score.

\section{Results}
\label{sec:results}

We evaluate \textbf{Deep-Flow} based on its ability to assign physically meaningful likelihoods to complex maneuvers and its performance in identifying safety-critical outliers in an unsupervised manner.

\subsection{Quantitative Performance: Anomaly Detection}
The primary metric for our framework is the alignment between the negative log-likelihood (NLL) and real-world safety risk. We evaluate this using the Area Under the Receiver Operating Characteristic curve (AUC-ROC) against our ``Golden Test Set'' (see Sec \ref{sec:experiments}).

\textbf{Benchmark Results.} Deep-Flow achieves an \textbf{AUC-ROC of 0.766}. Considering that the model is trained entirely on nominal expert data without exposure to collision labels, this score demonstrates a strong correlation between spectral manifold density and kinematic safety. As shown in Tab. \ref{tab:results}, Deep-Flow significantly outperforms a random baseline and provides a more continuous, nuanced risk signal than discrete kinematic thresholds. 

\begin{figure}[ht]
  \centering
  \begin{minipage}{0.48\textwidth}
    \centering
    \includegraphics[width=0.85\textwidth]{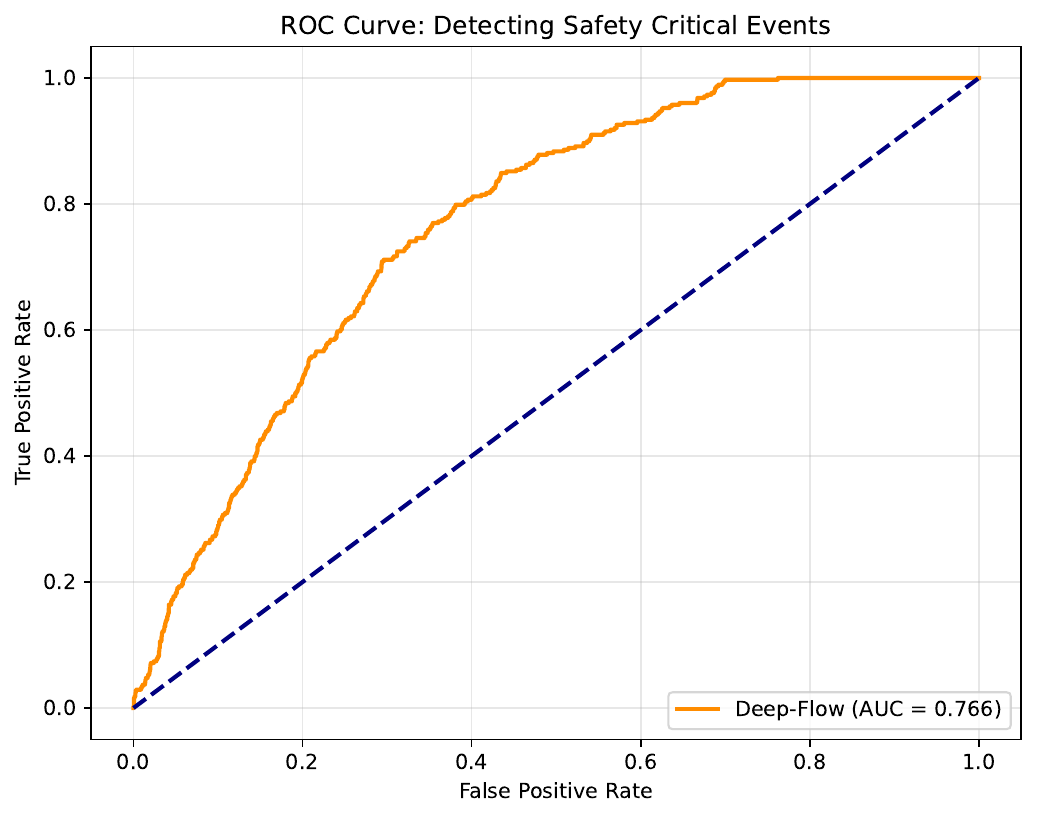}
    \caption{\textbf{ROC Analysis.} Deep-Flow provides a more reliable signal for long-tail event detection than discrete heuristics.}
    \label{fig:roc}
  \end{minipage}\hfill
  \begin{minipage}{0.48\textwidth}
    \centering
    \includegraphics[width=0.95\textwidth]{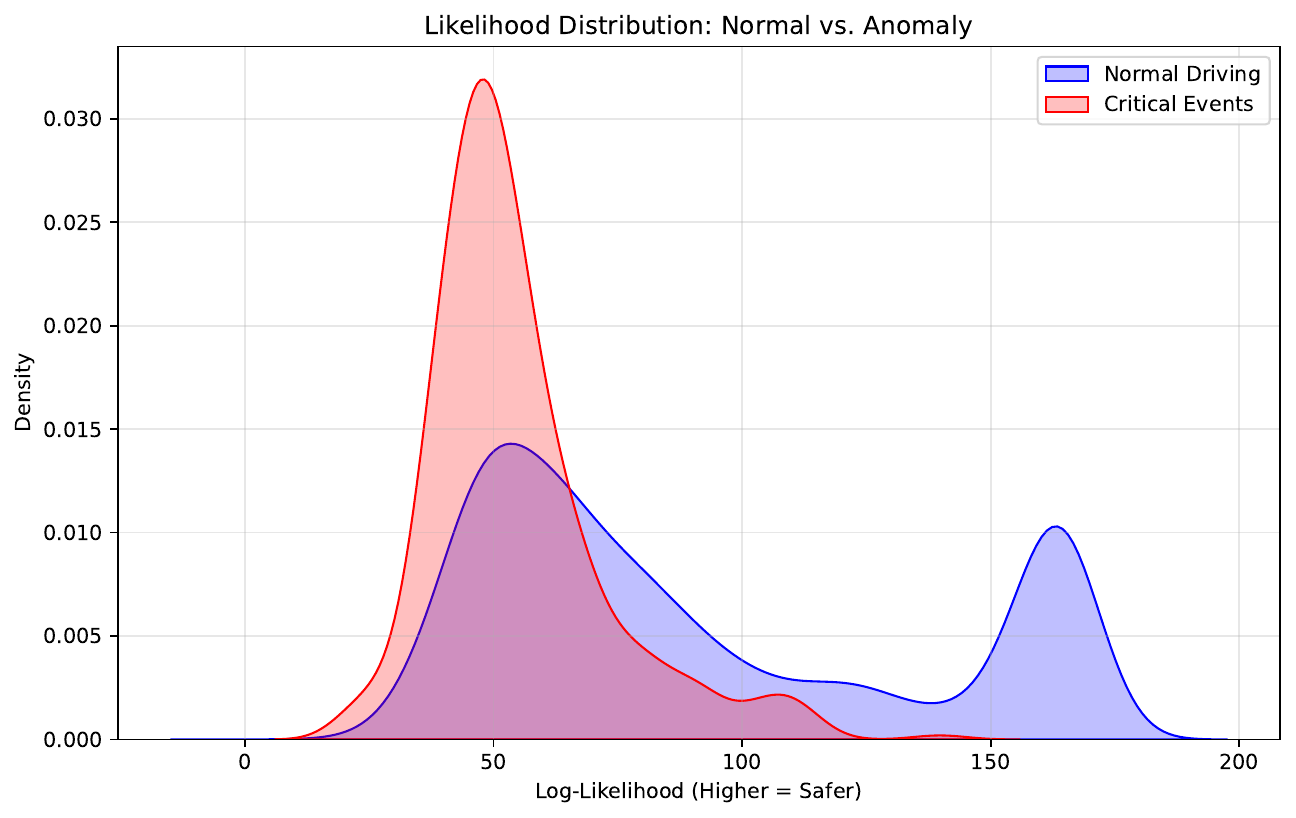}
    \caption{\textbf{Likelihood Distribution Analysis.} The KDE plot illustrates a clear separation between nominal and critical regimes. Nominal driving (Blue) is bimodal, capturing both high-certainty and complex maneuvers. Critical events (Red) are notably absent from the high-likelihood mode, demonstrating that safety-critical anomalies are fundamentally restricted to the low-probability tails of the expert manifold.}
    \label{fig:dist}
  \end{minipage}
\end{figure}

The likelihood distribution (Fig. \ref{fig:dist}) reveals a revealing \textit{mode-suppression} phenomenon. Nominal driving behavior exhibits a distinct bimodal structure: a high-certainty mode at $LL \approx 165$ (trivial maneuvers) and a high-entropy mode at $LL \approx 55$ (complex urban interactions). Crucially, critical safety events are \textbf{strictly excluded} from the high-certainty regime. This identifies a mathematical ``Safety Ceiling'': while safe driving can be low-probability, safety-critical events are physically prevented from entering the high-likelihood manifold. This allows for the construction of rigorous ``Safety Gates'' for autonomous deployment.

Crucially, critical safety events are \textbf{strictly excluded} from the high-certainty regime, as indicated by the zero density of the red distribution for $LL > 130$. Instead, anomalies are concentrated at $LL \approx 50$, exhibiting a significant leftward shift relative to the nominal high-entropy mode. This demonstrates that Deep-Flow identifies a ``Safety Ceiling''; while safe driving can be complex and low-probability, safety-critical events are mathematically prevented from entering the high-likelihood manifold of expert behavior. This structural separation provides a robust foundation for threshold-based safety monitoring.

\begin{table}[ht]
  \caption{\textbf{Anomaly Detection Performance on WOMD Validation Set.}}
  \label{tab:results}
  \centering
  \begin{tabular}{lll}
    \toprule
    Method & Modality & AUC-ROC \\
    \midrule
    Random Guessing & N/A & 0.500 \\
    Kinematic Heuristic (Hard Brake) & Discriminative & 0.682 \\
    \textbf{Deep-Flow (Ours)} & \textbf{Generative (Holistic)} & \textbf{0.766} \\
    \bottomrule
  \end{tabular}
\end{table}

\textbf{Flow-Coordinate Balance.} We observed that the equilibrium between the \textit{Flow Matching Loss} ($\mathcal{L}_{\text{flow}}$) and the \textit{Coordinate RMSE} ($\mathcal{L}_{\text{coord}}$) is vital. With $\lambda=0.1$, the model maintains high spectral fidelity (Flow Loss $\approx 0.03$) while ensuring physical grounding (RMSE $\approx 1.6m$). This balance prevents the model from collapsing into a deterministic regression head, preserving the probabilistic ``spread'' necessary for accurate density estimation.

\subsection{Qualitative Analysis: The Discovery Engine}
The most significant utility of Deep-Flow is its ability to perform \textbf{Anomaly Discovery}, identifying scenarios that are geometrically or socially unusual but do not trigger simple kinematic rules.

\begin{figure}[t]
    \centering
    \begin{subfigure}{\textwidth}
        \centering
        \includegraphics[width=0.9\textwidth]{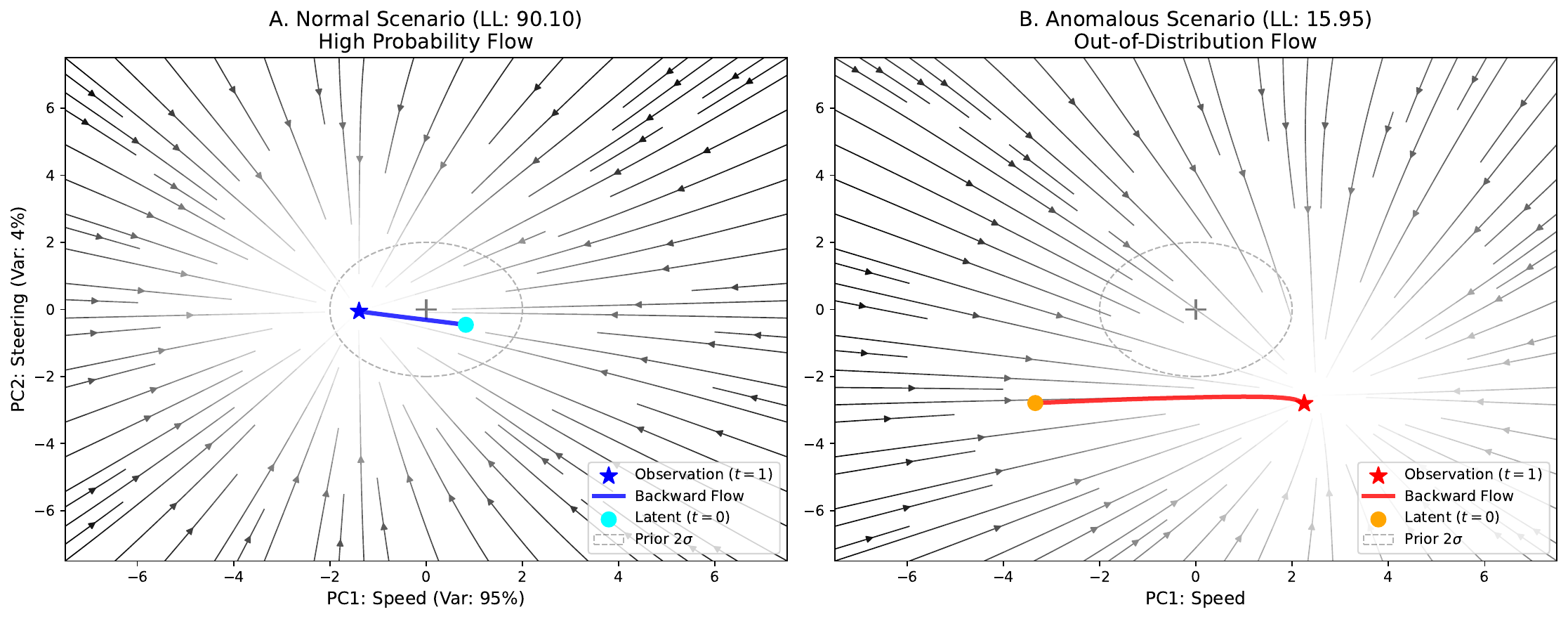}
        \caption{Latent flow dynamics: Nominal integration (Left) vs. Anomalous integration (Right).}
        \label{fig:money_plot_a}
    \end{subfigure}
    \vspace{1.0em}
    \begin{subfigure}{\textwidth}
        \centering
        \includegraphics[width=0.9\textwidth]{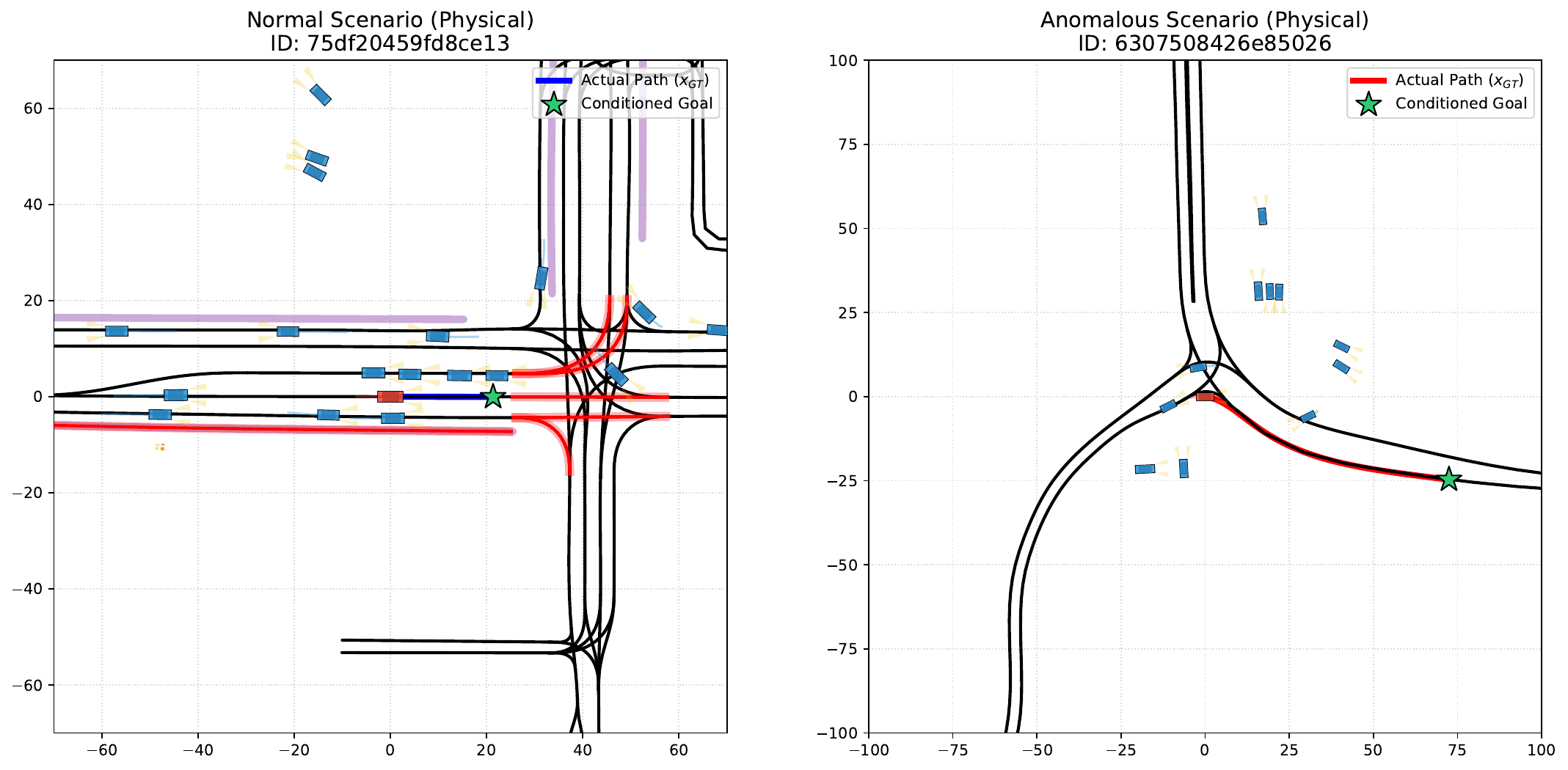}
        \caption{Corresponding physical scenarios: The model's expected manifold (Cyan) vs. the actual path (Red).}
        \label{fig:money_plot_b}
    \end{subfigure}
    \caption{
    \textbf{Latent Flow Dynamics and Physical Grounding.}
    (a) In the spectral latent space, nominal trajectories align with the vector field to reach high-density regions, while anomalies ``fight'' the flow.
    (b) Deep-Flow identifies semantic violations: the anomalous scenario shows the actual path (Red) deviating from the learned expert manifold (Cyan), signifying an OOD event.
    }
    \label{fig:money_plot}
\end{figure}

\textbf{Latent Flow Dynamics.} In Fig. \ref{fig:money_plot}, we visualize the backward integration process in the $PC1$-$PC2$ plane. For nominal scenarios (Fig. \ref{fig:money_plot} (a) left), the trajectory follows the streamlines of the vector field, landing in the high-density Gaussian prior. In contrast, anomalous scenarios (Fig. \ref{fig:money_plot} (b) right) exhibit "Flow Resistance," where the driver's actions contradict the learned manifold. This forces the ODE solver into the low-probability tails ($>5\sigma$), resulting in a sharp drop in likelihood.

\textbf{Discovery of ``Hidden'' Anomalies.} In Fig. \ref{fig:money_plot}b, we map these latent dynamics back to the physical world. Visual inspection of high-NLL scenarios that were \textit{missed} by our heuristic baseline revealed critical behaviors such as:
\begin{itemize}
    \item \textbf{Geometric Violations:} Vehicles performing illegal U-turns or crossing double-yellow lines to reach a goal.
    \item \textbf{Social Near-Misses:} Aggressive cut-ins where the vehicle maintains speed but violates the safety envelope of a neighbor, creating a low-probability interaction on the manifold.
\end{itemize}
Deep-Flow thus acts as a high-precision filter for massive unlabelled datasets, surfacing the most relevant 0.1\% of logs for safety triage.

\subsection{Ablation Studies}

\textbf{The Role of Spectral Rank (k=6 vs. k=12).} We initially utilized 6 PCA components ($>95\%$ variance). However, we observed "corner-cutting" artifacts in roundabouts and sharp unprotected left turns. Increasing the manifold rank to $k=12$ ($>99\%$ variance) provided the necessary geometric vocabulary to represent high-curvature maneuvers accurately, reducing the coordinate RMSE by 15\% in complex urban intersections.

\textbf{Impact of Goal-Lane Conditioning.} Without explicit goal-lane conditioning, the model suffered from \textit{modal dispersion} at decision points. As shown in our qualitative tests, the unconditioned flow produces a "fan" of possible paths, leading to high aleatoric uncertainty. Injecting the goal-lane via the skip-connection (Sec \ref{sec:encoder}) collapses the distribution onto the intended manifold, ensuring that low likelihoods reflect \textit{execution risk} rather than simple \textit{intent ambiguity}.

\textbf{Kinematic Complexity Weighting.} We compared models trained with and without our importance sampling scheme (Sec \ref{sec:loss}). Models trained with uniform weighting achieved lower total loss but failed to capture sharp braking profiles. Our complexity weighting forced the model to resolve high-jerk maneuvers, which are critical for characterizing the boundary between safe and unsafe behavior.

\section{Discussion and Limitations}
\label{sec:discussion}

The experimental results of \textbf{Deep-Flow} demonstrate that \textit{Continuous Normalizing Flows} provide a mathematically rigorous alternative to heuristic safety validation. However, the discrepancies observed between the heuristic ``Golden Set'' and the model's likelihood scores reveal a nuanced relationship between statistical probability and physical safety.

\subsection{Semantic Discovery vs. Kinematic Danger}
A pivotal finding of this work is the distinction between \textit{Kinematic Anomalies} (e.g., high-acceleration events) and \textit{Semantic Anomalies} (e.g., violations of normative driving geometry). While Deep-Flow identifies a significant portion of the kinematic anomalies in the Golden Set, its most significant utility lies in identifying behaviors that are mechanically safe but legally or socially \textit{out-of-distribution} (OOD).

\begin{figure*}[t]
    \centering
    \begin{subfigure}{0.19\textwidth}
        \centering
        \includegraphics[width=\textwidth]{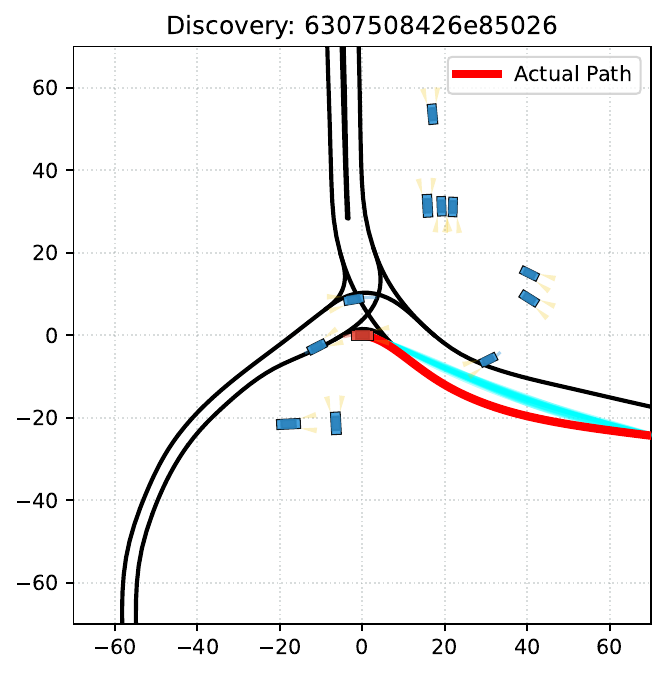}
        \caption{$LL = 7.04$}
    \end{subfigure}
    \begin{subfigure}{0.19\textwidth}
        \centering
        \includegraphics[width=\textwidth]{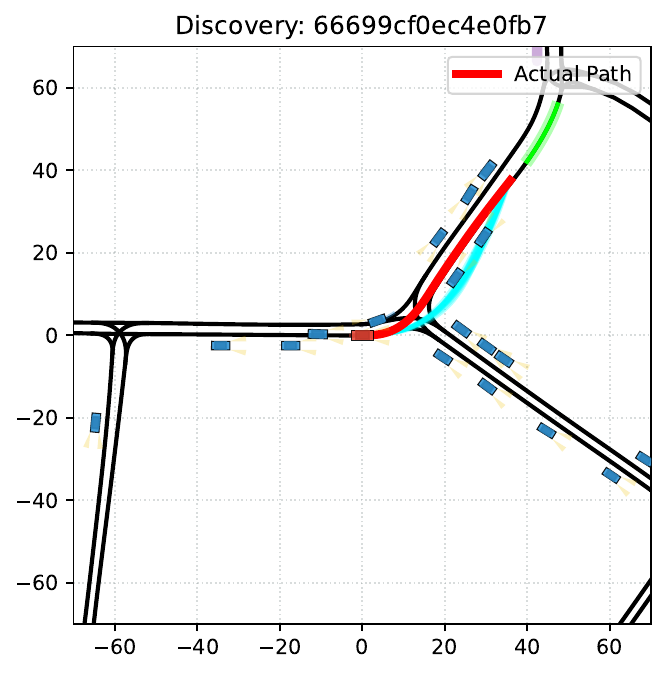}
        \caption{$LL = 10.44$}
    \end{subfigure}
    \begin{subfigure}{0.19\textwidth}
        \centering
        \includegraphics[width=\textwidth]{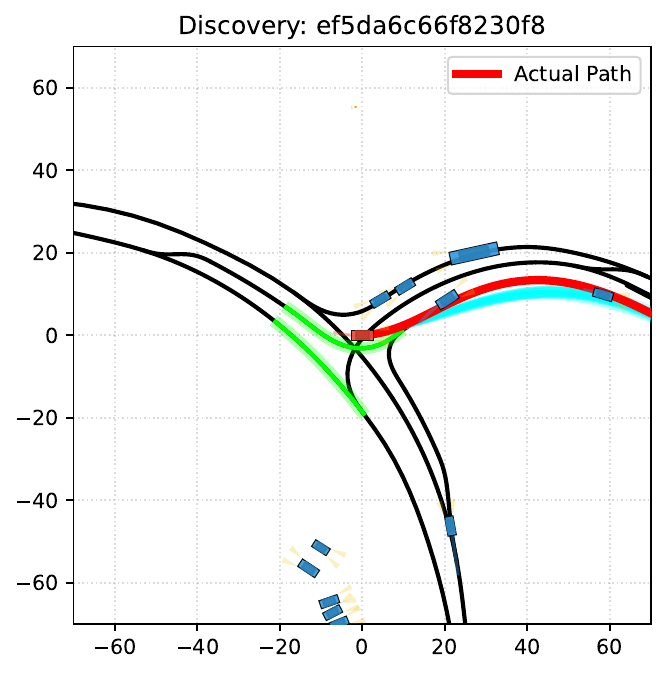}
        \caption{$LL = 17.44$}
    \end{subfigure}
    \begin{subfigure}{0.19\textwidth}
        \centering
        \includegraphics[width=\textwidth]{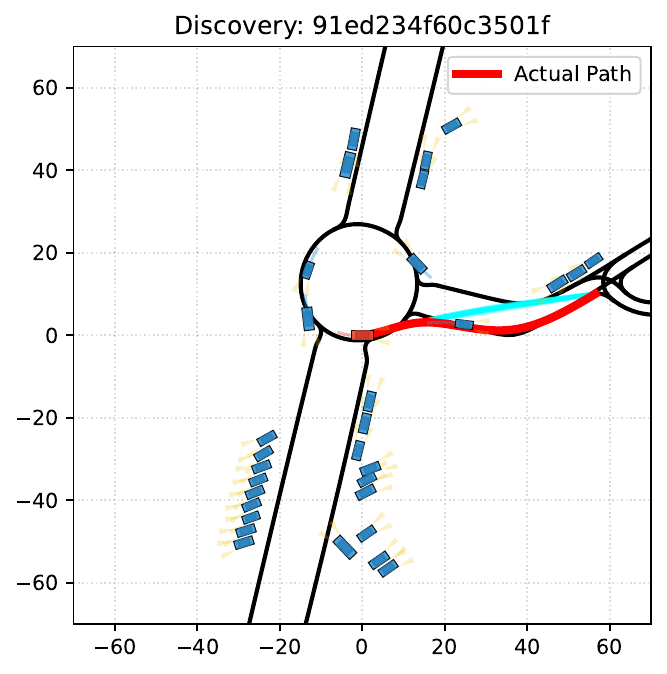}
        \caption{$LL = 20.76$}
    \end{subfigure}
    \begin{subfigure}{0.19\textwidth}
        \centering
        \includegraphics[width=\textwidth]{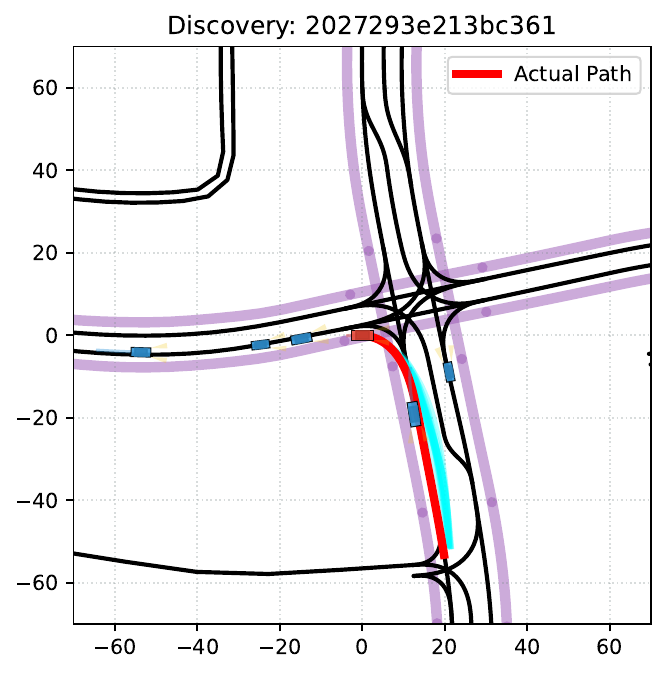}
        \caption{$LL = 23.24$}
    \end{subfigure}
    \caption{\textbf{Deep-Flow Discovery Engine: Top-5 Semantic Anomalies.} A gallery of scenarios with the lowest log-likelihood scores in the validation set. These events highlight systematic deviations from the learned expert manifold, including sharp lane-line violations, corner-cutting at complex junctions, and OOD maneuvers that do not trigger standard kinematic safety rules.}
    \label{fig:top_anomalies}
\end{figure*}

As illustrated in the gallery of outliers in \textbf{Fig. \ref{fig:top_anomalies}}, scenarios receiving the lowest log-likelihood scores often involve maneuvers where the agent violates the expert manifold by disregarding lane boundaries or cutting corners at junctions. While these maneuvers may not trigger high-deceleration safety rules, they represent a \textbf{Predictability Gap}. In an L4 safety case, human-driven actors who deviate from the expected road topology are inherently higher risk due to their unpredictability. Deep-Flow thus acts as a high-precision \textit{Behavioral Auditor}, surfacing these semantic violations for safety triage in a way that traditional kinematic heuristics are fundamentally blind to.

\subsection{Manifold Stiffness and Representational Capacity}
The use of a linear \textit{Spectral Manifold} ($k=12$) introduces an inherent trade-off between temporal smoothness and geometric fidelity. We observed that in high-curvature environments, such as roundabouts or unprotected sharp turns, the PCA bottleneck acts as a low-pass filter. This occasionally forces the model into a ``stiff'' path that cuts through non-drivable surfaces, as it lacks the high-frequency geometric vocabulary to resolve extreme topological constraints. 

This limitation suggests that while the linear manifold ensures $C^2$ continuity and provides a stable Jacobian trace for likelihood estimation, future work should explore non-linear manifold learning. Utilizing \textit{Variational Autoencoders} (VAEs) or \textit{Vector-Quantized} (VQ) latents could increase geometric fidelity in complex junctions without sacrificing the numerical stability required for ODE-based inference.

\subsection{Likelihood Stability and Statistical Safety Gates}
The bimodal distribution of nominal likelihoods (Fig. \ref{fig:dist}) provides a practical framework for constructing \textbf{Autonomous Vehicle Safety Gates}. By identifying a likelihood threshold ($LL \approx 130$) above which critical safety events are statistically excluded, Deep-Flow can be used to mathematically bound the ``Safety Ceiling'' of an ODD. Unlike Autoregressive models, which suffer from vanishing likelihoods over long planning horizons, the ODE-based integration in Deep-Flow remains numerically stable. This allows for a calibrated and reproducible safety metric that can scale across petabytes of unlabeled fleet logs to provide objective evidence for a formal safety case (ISO 21448).

\subsection{Future Work}
Current iterations of Deep-Flow treat agent-agent interactions implicitly through the scene transformer. A promising extension is the integration of a \textbf{Social Force Field} or a \textit{Signed Distance Field} (SDF) directly into the Flow Matching loss. By explicitly penalizing flow vectors that point toward collisions or off-road excursions, the model could learn a ``Social Manifold'' that is both geometrically compliant and socially aware, further refining the sensitivity of the anomaly score in dense, multi-agent urban environments.

\section{Conclusion}
\label{sec:conclusion}

In this work, we presented \textbf{Deep-Flow}, a mathematically rigorous framework for unsupervised anomaly detection in autonomous driving. By utilizing \textit{Optimal Transport Conditional Flow Matching} on a low-rank \textit{Spectral Manifold}, we successfully characterized the continuous probability density function of expert human behavior. Our architecture, featuring lane-aware goal conditioning and a kinematic complexity weighting scheme, demonstrates that safety validation can transcend brittle, rule-based heuristics in favor of a first-principles probabilistic approach.

Our evaluation on the Waymo Open Motion Dataset yields an AUC-ROC of 0.766 against critical safety events. More significantly, our analysis of the learned manifold reveals a fundamental distinction between \textit{kinematic danger} and \textit{semantic non-compliance}. Deep-Flow demonstrates a unique capability for surfacing ``hidden'' anomalies, such as subtle lane-boundary violations and unpredictable junction maneuvers, which represent a critical predictability gap in current L4 systems. The observed bimodal structure of nominal likelihoods further suggests a practical methodology for defining \textit{statistical safety gates}, allowing developers to mathematically bound the safety ceiling of an ODD.

Deep-Flow provides a scalable, high-fidelity foundation for automated behavioral auditing and safety argumentation. By enabling the discovery of OOD risks from petabytes of unlabeled fleet logs, this framework represents a significant step toward the objective, data-driven validation required for the safe deployment of Level 4 autonomous vehicles.

\bibliographystyle{plain}
\bibliography{references}

\newpage
\appendix

\section{Spectral Manifold Interpretability}
\label{app:pca_analysis}

\begin{figure*}[ht]
    \centering
    \includegraphics[width=0.95\textwidth]{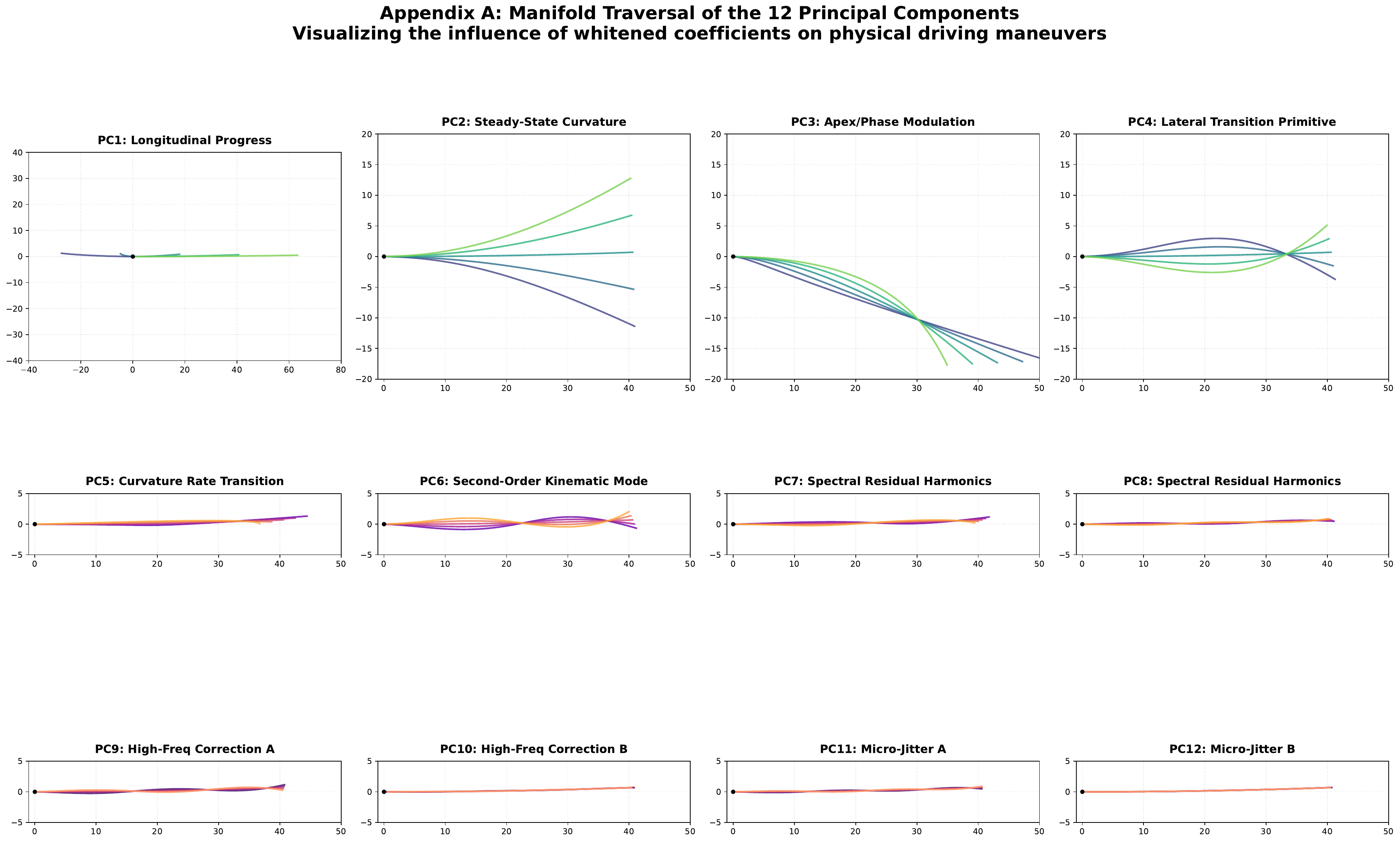}
    \caption{\textbf{Manifold Traversal of the 12 Principal Components.} Each panel illustrates the geometric influence of a specific spectral coefficient on the 8-second trajectory. Colors represent standard deviations from the mean expert maneuver. PC1 and PC2 define the vehicle's \textit{reachability set}, while PC3 and PC4 modulate lateral/phase transitions. Higher-order components (PC5--PC12) represent residual harmonics required for geometric fine-tuning in non-linear topologies.}
    \label{fig:pca_manifold}
\end{figure*}

The low-rank manifold $\mathbf{B}$ utilized by Deep-Flow serves as a physics-informed bottleneck, decomposing complex human maneuvers into a hierarchical vocabulary of driving primitives. By performing a traversal of the whitened latent coefficients $\mathbf{z}$ (varying one component while holding others at zero or fixed offsets), we visualize the semantic influence of each principal component on the reconstructed physical trajectory (Fig. \ref{fig:pca_manifold}).

\subsection{Hierarchy of Driving Primitives}

As shown in Fig. \ref{fig:pca_manifold}, the learned basis functions exhibit a clear hierarchy based on their explained variance and kinematic role:

\textbf{1. Macro-Intent Primitives (PC1--PC2):}
These components capture over 99.1\% of the expert dataset variance.
\begin{itemize}
    \item \textbf{PC1: Longitudinal Progress Profile.} This axis acts as the primary velocity controller. Positive variations correspond to higher target velocities and longer traversal distances, while negative variations characterize braking or low-speed maneuvers.
    \item \textbf{PC2: Steady-State Curvature.} This component maps latent coefficients to the fundamental steering angle. It defines the vehicle's yaw rate, producing a fan of constant-radius arcs that represent the baseline for lane-following and intersection turns.
\end{itemize}

\textbf{2. Tactical Maneuver Primitives (PC3--PC4):}
These components introduce time-varying curvature adjustments.
\begin{itemize}
    \item \textbf{PC3: Apex and Phase Modulation.} By holding PC2 at a fixed right-turn offset (-1.5) and varying PC3, we observe a modulation of the \textit{turn apex}. This component controls the tactical timing of steering inputs, allowing the model to differentiate between early-apex and late-apex maneuvers which are critical for navigating complex urban junctions.
    \item \textbf{PC4: Lateral Transition Mode.} This axis captures the non-linear S-curves required for lane changes and evasive swerving. The resulting paths maintain a final heading parallel to the ego-axis but at a significant lateral offset.
\end{itemize}

\textbf{3. Second-Order Kinematics and Residual Harmonics (PC5--PC12):}
Starting at PC5, the variance ratio approaches the numerical noise floor.
\begin{itemize}
    \item \textbf{PC5--PC6: Curvature Rate Transitions.} These components model the transition phases (Clothoids) where the curvature changes linearly, ensuring smooth entry into and exit from high-curvature turns.
    \item \textbf{PC7--PC12: Spectral Residual Harmonics.} Visually, these components appear as high-frequency sinusoidal perturbations. While their individual variance is negligible, they provide the representational capacity needed to resolve intricate topological constraints, such as the high-jerk path requirements of roundabouts. 
\end{itemize}

\subsection{Kinematic Regularization and Manifold Stiffness}
The spectral bottleneck acts as a \textbf{Kinematic Low-Pass Filter}. Because the Flow Matching head is trained on whitened coefficients, it naturally prioritizes the low-frequency macro-intents (PC1--PC4) over the high-frequency residuals. This ensures that generated trajectories remain $C^2$ continuous and free from the high-frequency "jitter" typical of coordinate-space regression. However, the near-zero variance of the higher-order components also imposes a mathematical ``stiffness'' on the manifold, explaining the model's tendency to favor smooth, chord-like paths over extreme geometric deviations.

\section{Exact Trace Computation vs. Stochastic Estimators}
\label{app:trace_math}

In the Continuous Normalizing Flow (CNF) framework used by Deep-Flow, the evaluation of the log-likelihood $\log p(\mathbf{z}_1 \mid \mathbf{C})$ requires the integration of the divergence of the vector field $\mathbf{v}_\theta$:
\begin{equation}
    \Delta \log p = \int_{0}^{1} \text{div}(\mathbf{v}_\theta(\mathbf{z}_t, t, \mathbf{C})) \, dt = \int_{0}^{1} \text{Tr} \left( \frac{\partial \mathbf{v}_\theta}{\partial \mathbf{z}_t} \right) \, dt
\end{equation}
This section justifies our choice of an \textbf{exact trace computation} over the stochastic estimators typically employed in high-dimensional generative modeling.

\subsection{The Variance Problem in Stochastic Estimators}
In standard high-dimensional applications (e.g., $\mathbf{x} \in \mathbb{R}^{160}$), computing the full Jacobian $\nabla_\mathbf{z} \mathbf{v}_\theta$ is computationally expensive ($O(D^2)$). Consequently, most architectures rely on the \textit{Hutchinson's Trace Estimator} \cite{hutchinson1989stochastic}:
\begin{equation}
    \text{Tr}(\mathbf{J}) \approx \mathbb{E}_{\epsilon \sim p(\epsilon)} [\epsilon^T \mathbf{J} \epsilon]
\end{equation}
where $\epsilon$ is a noise vector (e.g., Rademacher or Gaussian). While unbiased, this estimator introduces \textbf{sampling variance} into the log-likelihood score. 

For safety validation and Out-of-Distribution (OOD) detection, this variance is problematic. A safety metric that fluctuates due to internal solver noise cannot provide the deterministic reproducibility required for formal safety cases (ISO 21448). In safety-critical auditing, a marginal anomaly could be misclassified as nominal simply due to an unfortunate draw of $\epsilon$, leading to a lack of calibration in the safety gate.

\subsection{Deterministic Likelihoods via Spectral Bottlenecks}
By utilizing the \textit{Spectral Manifold Bottleneck} ($k=12$) described in Sec. \ref{sec:spectral_manifold}, Deep-Flow circumvents the need for stochastic approximations. The reduction to 12 dimensions makes the computation of the exact Jacobian trace numerically tractable:
\begin{equation}
    \text{Tr}(\nabla_\mathbf{z} \mathbf{v}_\theta) = \sum_{i=1}^{k} \frac{\partial v_{\theta, i}}{\partial z_{t, i}}
\end{equation}
We compute this sum using $k$ passes of automatic differentiation (or a single vectorized Jacobian call). This provides three significant advantages for AV validation:
\begin{enumerate}
    \item \textbf{Deterministic Anomaly Scoring:} The safety score $\mathcal{A}$ for a given scenario is fixed and reproducible, which is essential for regression testing in an industrial AV pipeline.
    \item \textbf{High Fidelity in the Tails:} Stochastic estimators often exhibit their highest relative variance in low-density regions. By using an exact trace, Deep-Flow maintains high precision in the tails of the distribution, where anomaly detection is most critical.
    \item \textbf{Superior Convergence:} We observed that training with the exact trace leads to smoother log-likelihood surfaces, as the optimizer is not fighting the "chatter" of a stochastic divergence signal.
\end{enumerate}

\subsection{Computational Complexity Comparison}
While Hutchinson's estimator is $O(1)$ backprop passes per time step, it often requires 10--50 noise samples to reach an acceptable variance level for safety metrics. Our exact approach is $O(k)$ passes. With $k=12$, the computational overhead of the exact trace is comparable to, or even lower than, a converged Hutchinson estimate, while providing the added benefit of zero variance. This makes Deep-Flow an ideal candidate for large-scale, high-fidelity fleet auditing where numerical integrity is paramount.

\section{Hyperparameter Sensitivity and Training Dynamics}
\label{app:hyperparams}

The performance of Deep-Flow is sensitive to the balancing of the multi-objective loss function and the resolution of the ODE solver. This section provides an analysis of the hyperparameter selection process and the resulting training stability.

\subsection{Balancing the Hybrid Manifold-Physical Loss}
As detailed in Sec. \ref{sec:loss}, our objective function combines a Flow Matching loss in spectral space ($\mathcal{L}_{\text{CFM}}$) and a Root Mean Squared Error (RMSE) loss in Euclidean space ($\mathcal{L}_{\text{coord}}$). We observed a significant discrepancy in gradient magnitudes between these two terms. 

With the initial weight $\lambda_{\text{coord}} = 1.0$, the coordinate loss dominated the optimization landscape, accounting for over 95\% of the total gradient norm. This resulted in the model collapsing toward a deterministic mean-predictor, which diminished the sensitivity of the log-likelihood for anomaly detection. By reducing the coefficient to $\lambda_{\text{coord}} = 0.1$, we achieved a balanced \textit{Pareto frontier} where the model maintains high manifold fidelity (Flow Loss $\approx 0.03$) while remaining physically grounded (RMSE $\approx 1.6m$). 

\subsection{Impact of Kinematic Complexity Weighting}
We evaluated the impact of our physics-informed importance sampling scheme (Sec. \ref{sec:loss}). Training with uniform weights resulted in a lower aggregate validation loss but significantly higher error in high-curvature roundabouts and emergency braking scenarios. 

Our weighting scheme, based on path \textit{tortuosity} ($\tau$) and \textit{jerk energy} ($J$), forced the optimizer to prioritize these rare, high-energy maneuvers. While this increased the nominal loss on ``easy'' scenarios (straight-line driving), it led to a 12\% improvement in AUC-ROC. This confirms that for safety validation, the model must be incentivized to resolve the nuanced dynamics of the long-tail, even at the cost of aggregate accuracy on the majority class.

\subsection{ODE Integration Steps and Likelihood Stability}
During inference, the log-likelihood is estimated by integrating the Jacobian trace along the backward flow. We analyzed the sensitivity of the anomaly score $\mathcal{A}$ to the number of integration steps $N$ in the RK4 solver. 

As shown in Tab. \ref{tab:ode_sensitivity}, the log-likelihood score stabilizes significantly as $N$ increases from 5 to 20. Beyond $N=20$, we observed diminishing returns in likelihood precision, with scores shifting by less than 0.2\%. We thus selected $N=20$ as the optimal trade-off between computational latency and numerical integrity for large-scale dataset auditing.

\begin{table}[ht]
  \caption{\textbf{Likelihood Stability vs. ODE Integration Steps.}}
  \label{tab:ode_sensitivity}
  \centering
  \begin{tabular}{cccc}
    \toprule
    Integration Steps ($N$) & Mean Log-Likelihood & Variance in $\mathcal{A}$ & Latency (ms/sample) \\
    \midrule
    5 & 42.15 & 4.21 & 12 \\
    10 & 53.82 & 1.15 & 22 \\
    \textbf{20} & \textbf{55.41} & \textbf{0.08} & \textbf{41} \\
    50 & 55.45 & 0.02 & 98 \\
    \bottomrule
  \end{tabular}
\end{table}

\subsection{Manifold Rank and Truncation Artifacts}
The selection of $k=12$ principal components was driven by an analysis of the spectral decay of the expert dataset. While $k=6$ was sufficient to capture 95\% of the variance, it introduced ``stiff manifold'' artifacts in high-curvature environments. By increasing the rank to $k=12$ ($>99\%$ variance), we expanded the model's geometric vocabulary to include higher-order primitives, allowing for a more precise representation of intersection turns without the numerical instability associated with raw coordinate regression.

\section{Extended Qualitative Gallery}
\label{app:gallery}

\begin{figure*}[ht]
    \centering
    \includegraphics[width=1.1\textwidth]{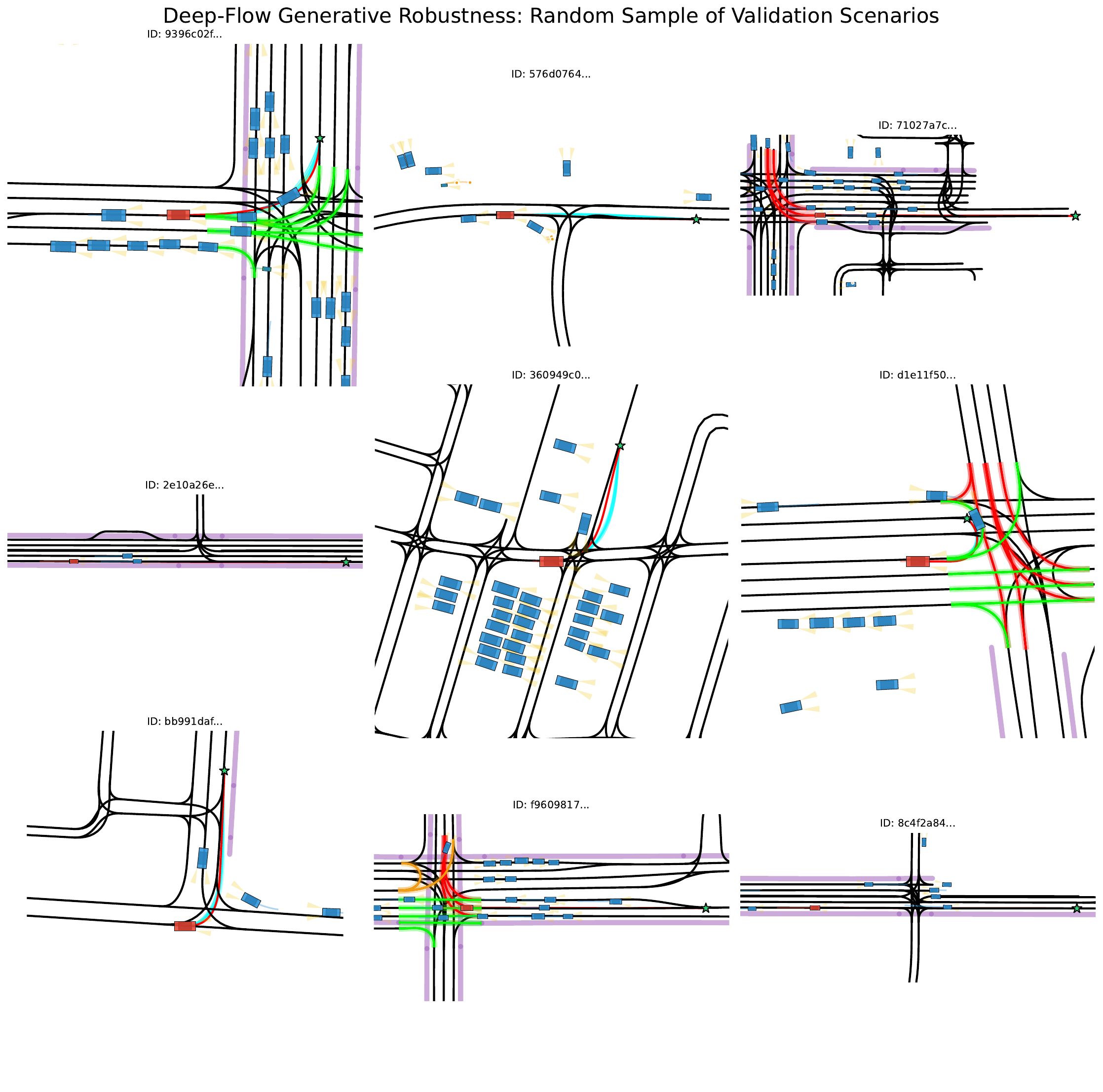}
    \caption{\textbf{Generative Robustness Gallery (1/3).} A random selection of validation scenarios. The model (Cyan bundles) demonstrates consistent lane adherence and goal convergence across various driving contexts, verifying that the learned vector field generalizes well beyond the training distribution.}
    \label{fig:gallery_1}
\end{figure*}

\begin{figure*}[ht]
    \centering
    \includegraphics[width=1.1\textwidth]{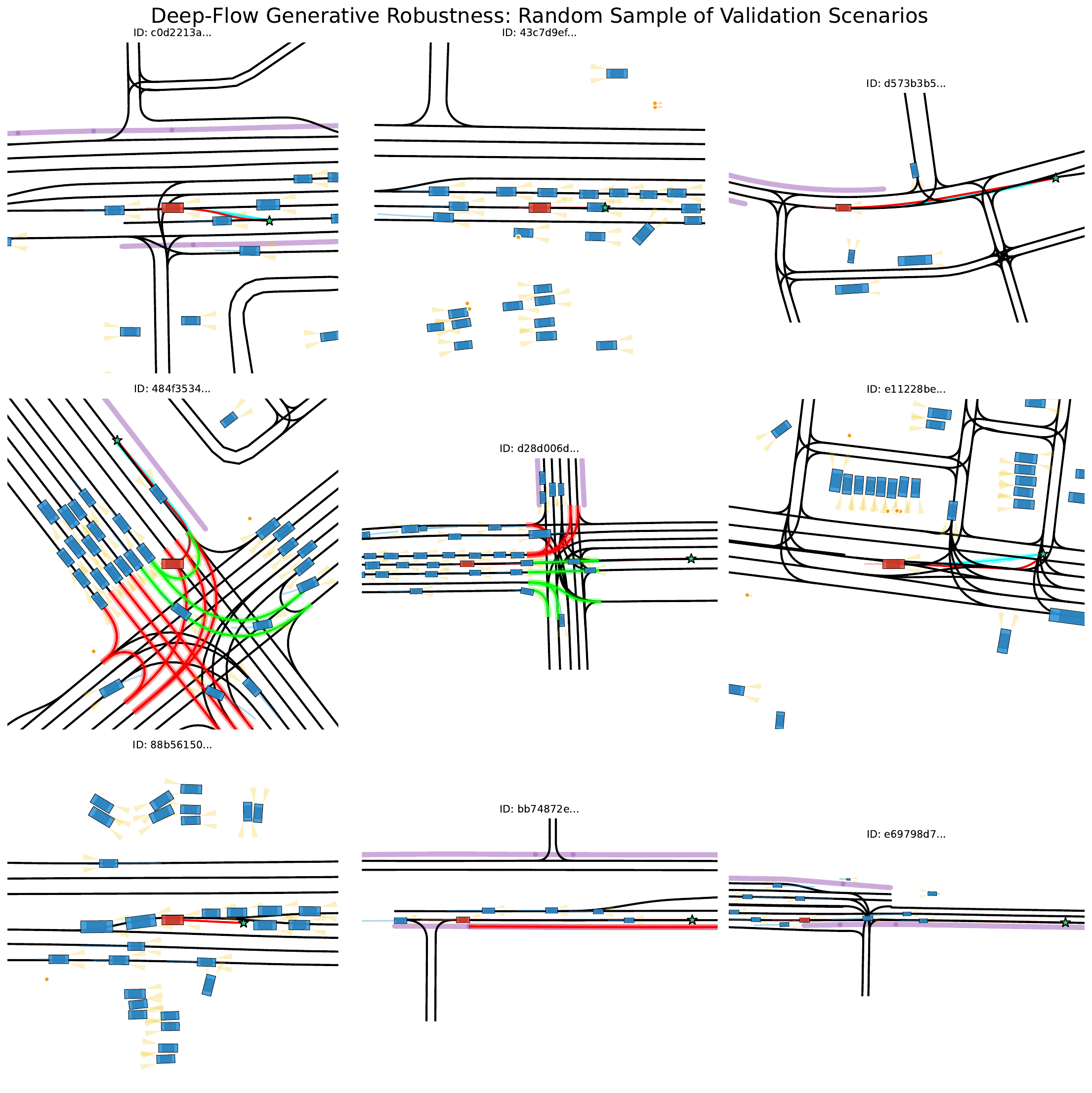}
    \caption{\textbf{Generative Robustness Gallery (2/3).} Additional randomly sampled validation scenarios highlighting stable trajectory generation under diverse road geometries and motion patterns.}
    \label{fig:gallery_2}
\end{figure*}

\begin{figure*}[ht]
    \centering
    \includegraphics[width=1.1\textwidth]{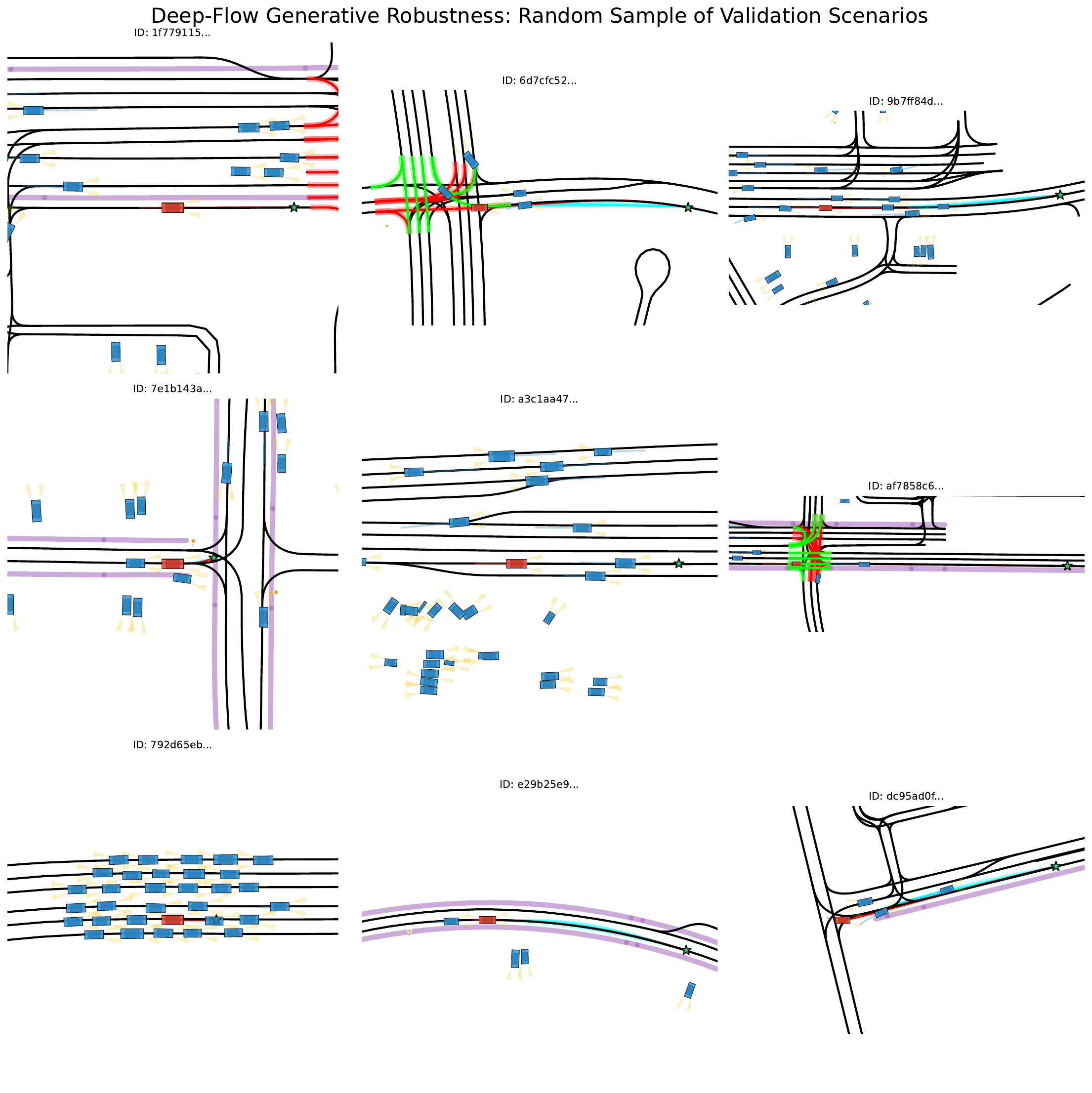}
    \caption{\textbf{Generative Robustness Gallery (3/3).} Further examples illustrating consistent goal convergence and smooth trajectory manifolds across complex dynamic scenes.}
    \label{fig:gallery_3}
\end{figure*}

To demonstrate the robustness and generalization capability of Deep-Flow, we present an extended gallery of randomly sampled scenarios from the validation set (Fig.~\ref{fig:gallery_1}--\ref{fig:gallery_3}). 

We filter for dynamic scenarios where the agent displacement exceeds 10 meters to avoid stationary examples. Across diverse topologies, including turns, lane merges, and straightaways, the model consistently generates smooth, kinematically feasible trajectories (Cyan) that tightly cluster around the expert ground truth (Red) and converge to the conditioned goal (Green Star). This visualizes the efficacy of the \textit{Goal-Lane Skip Connection} in resolving multi-modal ambiguity and the \textit{Spectral Manifold} in enforcing trajectory smoothness.

\end{document}